%% file: tech_report.tex
\pdfoutput=1

\documentclass[10pt,twocolumn,letterpaper,pagenumbers]{article}

\usepackage[]{cvpr} 
\usepackage[accsupp]{axessibility}  

\usepackage[utf8]{inputenc} 
\usepackage[T1]{fontenc}    
\usepackage{hyperref}       
\usepackage{url}            
\usepackage{booktabs}       
\usepackage{amsfonts}       
\usepackage{nicefrac}       
\usepackage{microtype}      
\usepackage{xcolor}         

\usepackage{graphicx}
\usepackage{amsmath}
\usepackage{amssymb}
\usepackage{booktabs}
\usepackage{float}

\usepackage{bm}
\usepackage{bbm}
\usepackage{mathtools}
\usepackage{amsfonts}       
\usepackage{nicefrac}       
\usepackage{microtype}      
\usepackage{multirow}
\usepackage{adjustbox}
\usepackage{wrapfig}
\usepackage{placeins}
\usepackage[normalem]{ulem}
\useunder{\uline}{\ul}{}

\usepackage{color}
\definecolor{Red7}{rgb}{0.941, 0.243, 0.243}
\definecolor{Green7}{RGB}{55, 178, 77}
\definecolor{Tdgreen}{rgb}{0,0.4,0.7}
\usepackage{pifont}
\usepackage[capitalize]{cleveref}
\crefname{section}{Sec.}{Secs.}
\Crefname{section}{Section}{Sections}
\Crefname{table}{Table}{Tables}
\crefname{table}{Tab.}{Tabs.}

\newfloat{figtab}{htb}{fgtb}
\makeatletter
  \newcommand\figcaption{\def\@captype{figure}\caption}
  \newcommand\tabcaption{\def\@captype{table}\caption}
\makeatother

\title{2nd Place Winning Solution for the CVPR2023 Visual Anomaly and Novelty Detection Challenge: Multimodal Prompting for Data-centric  Anomaly Detection}

\author{
  Yunkang Cao$^{1}$\footnotemark[1]\quad 
  Xiaohao Xu$^{1}$\footnotemark[1]\quad 
  Chen Sun$^{1}$ \quad 
  {Yuqi Cheng}$^{1}$ \\
  {Liang Gao}$^{1}$  \quad
  {Weiming Shen$^{1}$\footnotemark[4]} 
  \\
$^1$ State Key Laboratory of Digital Manufacturing 
Equipment and Technology,\\ Huazhong University of Science and Technology, China\\
  \texttt{\{cyk\_hust, sun\_chen, chengyuqi, gaoliang\}@hust.edu.cn}  \\  \texttt{xxh11102019@outlook.com},  \texttt{wshen@ieee.org} \\
}

\begin{document}

\maketitle
\renewcommand{\thefootnote}{\fnsymbol{footnote}}
\footnotetext[1]{Equal Contribution.}
\footnotetext[4]{Corresponding Author.}

\renewcommand{\thefootnote}{\arabic{footnote}}

\begin{abstract}
This technical report introduces the winning solution of the team \textit{Segment Any Anomaly} for the CVPR2023 Visual Anomaly and Novelty Detection (VAND) challenge. Going beyond uni-modal prompt, \textit{e.g.}, language prompt, we present a novel framework, \textit{i.e.}, Segment Any Anomaly + (SAA$+$), for zero-shot anomaly segmentation with multi-modal prompts for the regularization of cascaded modern foundation models. Inspired by the great zero-shot generalization ability of foundation models like Segment Anything, we first explore their assembly (SAA) to leverage diverse multi-modal prior knowledge for anomaly localization. Subsequently,  we further introduce multimodal prompts (SAA$+$) derived from domain expert knowledge and target image context to enable the non-parameter adaptation of foundation models to anomaly segmentation. The proposed SAA$+$ model achieves state-of-the-art performance on several anomaly segmentation benchmarks, including VisA and MVTec-AD, in the zero-shot setting. We will release the code of our winning solution for the CVPR2023 VAND challenge at \href{Segment-Any-Anomaly}{https://github.com/caoyunkang/Segment-Any-Anomaly} \footnote{The extended-version paper with more details is available at ~\cite{cao2023segment}.}

\end{abstract}

\section{Introduction}

\label{sec:intro}

\begin{figure}[t]
  \centering
  \includegraphics[width=\linewidth]{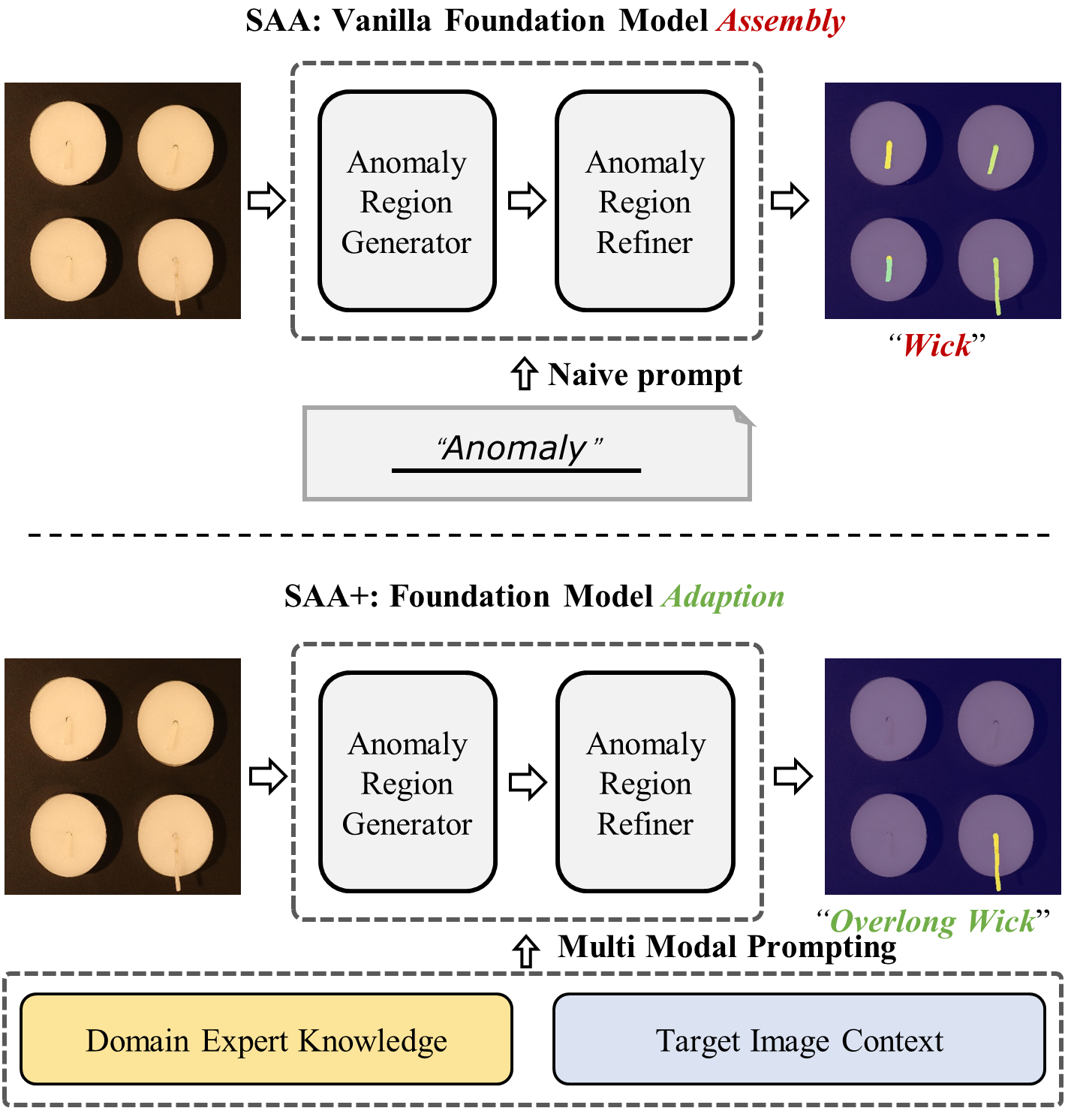}
   \vspace{-5mm}
   \caption{Towards segmenting any anomaly without training, we first construct a vanilla baseline (Segment Any Anomaly, SAA) by prompting into a cascade of anomaly region generator (\textit{e.g.}, a prompt-guided object detection foundation model~\cite{liu2023grounding}) and  anomaly region refiner (\textit{e.g.}, a segmentation foundation model~\cite{kirillov2023segment}) modules via a naive class-agnostic language prompt (\textit{e.g.}, ``Anomaly''). However, SAA shows the severe false-alarm problem, which falsely detects all the ``\texttt{\textcolor[RGB]{207,63,63}{wick}}'' rather than the ground-truth anomaly region (the ``\texttt{\textcolor[RGB]{112,173,71}{overlong wick}}''). Thus, we further strengthen the regularization of foundation models via multimodal prompts in the revamped model (Segment Any Anomaly +, SAA$+$), which successfully helps identify the anomaly region.}
  \label{fig:teaser}
  \vspace{-1mm}
\end{figure}

 Anomaly segmentation~\cite{deng2022anomaly,cao2022informative, wan_position_2022, cao_collaborative_2023} have gained great popularity in industrial quality control~\cite{bergmann2019mvtec}, medical diagnoses~\cite{baur_autoencoders_2021}, etc. We focus on the setting of zero-shot anomaly segmentation (ZSAS) on images, which aims at utilizing neither normal nor abnormal samples for segmenting any anomalies in countless objects.

Recently, foundation models, \textit{e.g.}, SAM~\cite{kirillov2023segment} and CLIP~\cite{radford2021learning}, exhibit great zero-shot visual perception abilities by retrieving prior knowledge stored in these models via prompting~\cite{ju_prompting_2022, jia_visual_2022, zang_unified_2022, shen_multitask_2022, zhou_learning_2022}.  In this work, we first construct a vanilla baseline, \textit{i.e.}, Segment Any Anomaly (SAA), by cascading prompt-guided object detection~\cite{liu2023grounding} and segmentation foundation models~\cite{kirillov2023segment}, which serve as Anomaly Region Generator and Anomaly Region Refiner, respectively. Following the practice to unlock foundation model knowledge~\cite{clipseg2022,jeong2023winclip}, naive language prompts, \textit{e.g.}, ``\verb|defect|'' or ``\verb|anomaly|'', are utilized to segment desired anomalies for a target image. In specific, the language prompt is used to prompt the Anomaly Region Generator to generate prompt-conditioned box-level regions for desired anomaly regions. Then these regions are refined in the Anomaly Region Refiner to produce final predictions, \textit{i.e.}, masks, for anomaly segmentation. 

However, as is shown in Figure \ref{fig:teaser}, vanilla foundation model assembly (SAA) tends to cause significant false alarms, \textit{e.g.}, SAA wrongly refers to all wicks as anomalies whereas only the overlong wick is a real anomaly, which we attribute to the \textit{ambiguity} brought by naive language prompts. Firstly, conventional language prompts may become ineffective when facing the domain shift between pre-training data distribution of foundation models and downstream datasets for anomaly segmentation. Secondly, the degree of ``\verb|anomaly|'' for a target depends on the object context, which is hard for coarse-grained language prompts, \textit{e.g.}, ``\verb|an anomaly region|'', to express exactly.

Thus, to reduce the language ambiguity, we incorporate domain expert knowledge and target image context in our revamped framework, \textit{i.e.}, Segment Any Anomaly + (SAA$+$).
Specifically, expert knowledge provides detailed descriptions of anomalies that are relevant to the target in open-world scenarios. We utilize more specific descriptions as in-context prompts, effectively aligning the image content in both pre-trained and target datasets. Besides, we utilize target image context to reliably identify and adaptively calibrate anomaly segmentation predictions~\cite{object_calibration, xu2022reliable}. By leveraging the rich contextual information present in the target image, we can accurately associate the object context with the final anomaly predictions.

\begin{figure*}[!t]
  \centering
  \includegraphics[width=\linewidth]{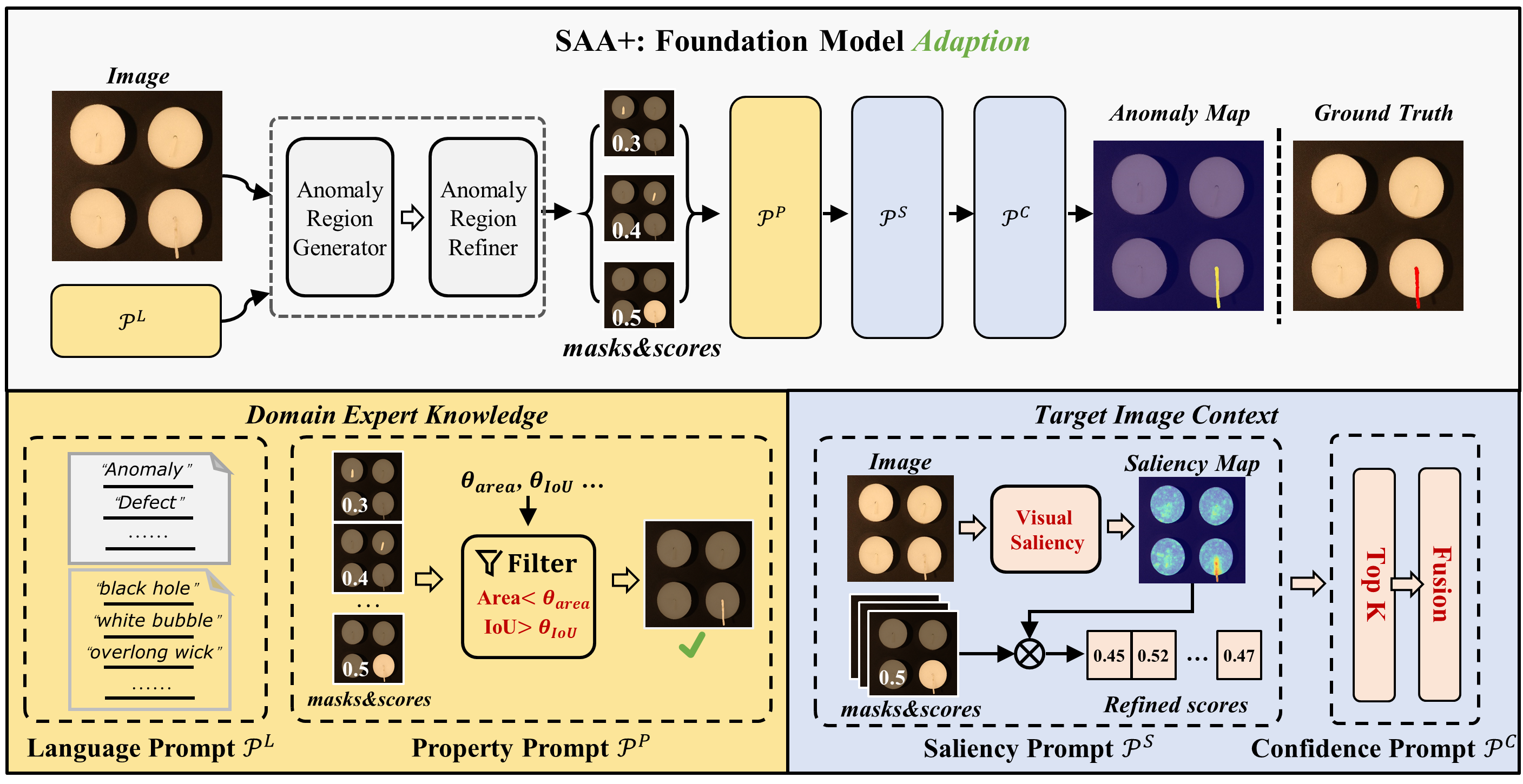}
  \vspace{-0.15in}
  \caption{\textbf{Overview of the proposed Segment Any Anomaly + (SAA+) framework.} We adapt foundation models to zero-shot anomaly segmentation via multimodal prompt regularization. In specific, apart from naive class-agnostic language prompts, the regularization comes from both domain expert knowledge, including more detailed class-specific language and object property prompts, and target image context, including visual saliency and confidence ranking-related prompts.
  }  
  \label{fig:framework}
  
\end{figure*}

\section{Starting from Vanilla Foundation Model Assembly with Language Prompt}

\subsection{Problem Definition: Zero-shot Anomaly Segmentation (ZSAS)}

The goal of ZSAS is to perform anomaly segmentation on new objects without requiring any corresponding object training data. ZSAS seeks to create an anomaly map $\mathbf{A} \in [0,1]^{h \times w \times 1}$ based on an empty training set $\emptyset$, in order to identify the anomaly degree for individual pixels in an image $\mathbf{I} \in \mathbb{R}^{h \times w \times 3}$ that includes novel objects. The ZSAS task has the potential to significantly reduce data requirements and lower real-world inspection deployment costs.

\subsection{Baseline: Segment Any Anomaly (SAA) }

For ZSAS, we start by constructing a vanilla foundation model assembly, \textit{i.e.}, Segment Any Anomaly (SAA), as shown in Fig. \ref{fig:teaser}, which consists of an Anomaly Region Generator and an Anomaly Region Refiner.

\subsubsection{Anomaly Region Generator}
There we base the architecture of the region detector on a text-guided open-set object detection architecture for visual grounding. Specifically, given the bounding-box-level region set $\mathcal{R}^B$, and their corresponding confidence score set $\mathcal{S}$, the module of anomaly region generator ($\mathrm{Generator}$) can be formulated as,
\begin{equation}
\label{eq:det}
    \mathcal{R}^B, \mathcal{S} := \mathrm{Generator}(\mathbf{I},\mathcal{T})
\end{equation}

\subsubsection{Anomaly Region Refiner}

To generate pixel-wise anomaly segmentation results, we propose Anomaly Region Refiner to refine the bounding-box-level anomaly region candidates into an anomaly segmentation mask set through SAM~\cite{kirillov2023segment}. SAM accepts the bounding box candidates $\mathcal{R}^B$ as prompts and obtain pixel-wise segmentation masks $\mathcal{R}$. The module of the Anomaly Region Refiner ($\mathrm{Refiner}$) can be formulated as follows,
\begin{equation}
    \mathcal{R} := \mathrm{Refiner}(\mathbf{I}, \mathcal{R}^B)
\end{equation}

Till then, we obtain the set of regions in the form of high-quality segmentation masks $\mathcal{R}$ with corresponding confidence scores $\mathcal{S}$. We summarize the framework ($\mathrm{SAA}$) as follows,
\begin{equation}
    \mathcal{R}, \mathcal{S} := \mathrm{SAA}(\mathbf{I}, \mathcal{T}_n)
\end{equation}
where $\mathcal{T}_n$ is a naive class-agnostic language prompt, \textit{e.g.}, ``\verb|anomaly|'', utilized in SAA. 

\subsection{Observation: Vanilla Language Prompt Fails to Unleash the Power of Foundation Models}

We present some preliminary experiments to evaluate the efficacy of vanilla foundation model assembly for ZSAS. Despite the simplicity and intuitiveness of the solution, we observe a \textit{language ambiguity} issue. Specifically, certain language prompts, such as ``\verb|anomaly|'', may fail to detect the desired anomaly regions. For instance, as depicted in Fig. \ref{fig:teaser}, all ``\verb|wick|'' is erroneously identified as an anomaly by the SAA with the ``\verb|anomaly|'' prompt.  We propose introducing multimodal prompts generated by domain expert knowledge and the target image context to reduce language ambiguity, thereby achieving better ZSAS performance. 

\section{Adapting Foundation Models to Anomaly Segmentation with Multi-modal Prompts}
\label{sec:method}  

To address language ambiguity in SAA and improve its ability on ZSAS, we propose an upgraded version called SAA$+$, incorporating multimodal prompts, as shown in Fig. \ref{fig:framework}. In addition to leveraging the knowledge gained from pre-trained foundation models, SAA$+$ utilizes both domain expert knowledge and target image context to generate more accurate anomaly region masks. We provide further details on these multimodal prompts below.

\subsection{Prompts Generated from Domain Expert Knowledge}

To address language ambiguity, we leverage domain expert knowledge that contains useful prior information about the target anomaly regions. Specifically, although experts may not provide a comprehensive list of potential open-world anomalies for a new product, they can identify some candidates based on their past experiences with similar products. Domain expert knowledge enables us to refine the naive ``\verb|anomaly|'' prompt into more specific prompts that describe the anomaly state in greater detail. In addition to language prompts, we introduce property prompts to complement the lack of awareness on specific properties like ``\verb|count|'' and ``\verb|area|'' ~\cite{paiss_count_2023} in existing foundation models~\cite{paiss_count_2023}.

\subsubsection{Anomaly Language Expression as Prompt}

To describe potential open-world anomalies, we propose designing more precise language prompts. These prompts are categorized into two types: class-agnostic and class-specific prompts. 

\noindent \textbf{Class-agnostic prompts ($\mathcal{T}_{\rm a}$}) are general prompts that describe anomalies that are not specific to any particular category, \textit{e.g.}, ``\verb|anomaly|'' and ``\verb|defect|''.

\noindent \textbf{Class-specific prompts ($\mathcal{T}_{\rm s}$}) are designed based on expert knowledge of abnormal patterns with similar products to supplement more specific anomaly details.
We use prompts already employed in the pre-trained visual-linguistic dataset, \textit{e.g.}, ``\verb|black hole|'' and ``\verb|white bubble|'', to query the desired regions. This approach reformulates the task of finding an anomaly region into locating objects with a specific anomaly state expression, which is more straightforward to utilize foundation models than identifying ``\verb|anomaly|'' within an object context.

By prompting SAA with anomaly language prompts $\mathcal{P}^L=\{ \mathcal{T}_{\rm a}, \mathcal{T}_ {\rm s}   \}$ derived from domain expert knowledge, we generate finer anomaly region candidates $\mathcal{R}$ and corresponding confidence scores $\mathcal{S}$.

\subsubsection{Anomaly Object Property as Prompt}

Current foundation models~\cite{liu2023grounding} have limitations when it comes to referring to objects with specific property descriptions, such as size or location~\cite{paiss_count_2023}, which are important for describing anomalies, such as ``\verb|The small black hole on the left.|'' To incorporate this critical expert knowledge, we propose using anomaly property prompts formulated as rules rather than language. Specifically, we consider the location and area of anomalies.

\noindent\textbf{Anomaly Location.} Anomalies typically locate within the inspected objects. To guarantee this, we calculate the intersection over union (IoU) between the potential anomaly regions and the inspected object. By applying an expert-derived IoU threshold, denoted as $\theta_{IoU}$, we filter out anomaly candidates with IoU values below this threshold, retaining regions that are more likely to be abnormal.

\noindent\textbf{Anomaly Area.} The size of an anomaly, as reflected by its area, is also a property that can provide useful information. In general, anomalies should be smaller than the size of the inspected object. Experts can provide a suitable threshold value $\theta_{area}$ for the specific type of anomaly being considered. Candidates with areas unmatched with $\theta_{area} \cdot \mathrm{Object Area}$ can then be filtered out.

By combining the two property prompts $\mathcal{P}^P=\{ \theta_{area}, \theta_{IoU} \}$, we can filter the set of candidate regions $\mathcal{R}$ to obtain a subset of selected candidates $\mathcal{R}^P$ with corresponding confidence scores $\mathcal{S}^P$ using the filter function ($\mathrm{Filter}$),
\begin{equation}
    \mathcal{R}^P, \mathcal{S}^P := \mathrm{Filter}(\mathcal{R}, \mathcal{P}^P)
\end{equation}

\subsection{Prompts Derived from Target Image Context}

Besides incorporating domain expert knowledge, we can leverage the information provided by the input image itself to improve the accuracy of anomaly region detection. In this regard, we propose two prompts induced by image context.

\subsubsection{Anomaly Saliency as Prompt}
Predictions generated by foundation models like~\cite{liu2023grounding} using the prompt ``\verb|defect|'' can be unreliable due to the domain gap between pre-trained language-vision datasets~\cite{Laion400} and targeted anomaly segmentation datasets~\cite{bergmann2019mvtec,zou2022spot}. To calibrate the confidence scores of individual predictions, we propose Anomaly Saliency Prompt mimicking human intuition. In specific, humans can recognize anomaly regions by their discrepancy with their surrounding regions \cite{aota2023zero}, \textit{i.e.}, visual saliency could indicate the anomaly degree. Hence, we calculate a saliency map ($\mathbf{s}$) for the input image by computing the average distances between the corresponding pixel feature ($\mathbf{f}$) and its $N$ nearest neighbors, 
\begin{equation}
\label{eq:saliency-map}
    \mathbf{s}_{ij} := \frac{1}{N}\sum\limits_{\mathbf{f}\in N_p(\mathbf{f}_{ij})}(1-  \langle \mathbf{f}_{ij},\mathbf{f}  \rangle)
\end{equation}
where $(i,j)$ denotes to the pixel location, $N_p(\mathbf{f}_{ij})$ denotes to the $N$ nearest neighbors of the corresponding pixel, and $\langle \cdot, \cdot \rangle$ refers to the cosine similarity. We use pre-trained CNNs from large-scale image datasets~\cite{hinton2012imagenet} to extract image features, ensuring the descriptiveness of features. The saliency map indicates how different a region is from other regions. The saliency prompts $\mathcal{P}^{S}$ are defined as the exponential average saliency value within the corresponding region masks, 
\begin{equation}
\label{eq:score_saliency}
    \mathcal{P}^{S} := \left\{ \exp(\frac{\sum_{i j}\mathbf{r}_{i j}\mathbf{s}_{i j}}{\sum_{i j}\mathbf{r}_{i j}}) \quad | \quad \mathbf{r} \in \mathcal{R}^P \right\}
\end{equation}

The saliency prompts provide reliable indications of the confidence of anomaly regions. These prompts are employed to recalibrate the confidence scores generated by the foundation models, yielding new rescaled scores $\mathcal{S}^{S}$ based on the anomaly saliency prompts $\mathcal{P}^{S}$. These rescaled scores provide a combined measure that takes into account both the confidence derived from the foundation models and the saliency of the region candidate. The process is formulated as follows,
\begin{equation}
\label{eq:rescore}
    \mathcal{S}^{S} := \left\{ p \cdot s  \quad | \quad   p \in \mathcal{P}^{S},   s \in \mathcal{S}^P \right\}
\end{equation}

\subsubsection{Anomaly Confidence as Prompt}

Typically, the number of anomaly regions in an inspected object is limited. Therefore, we propose anomaly confidence prompts $\mathcal{P}^C$ to identify the $K$ candidates with the highest confidence scores based on the image content and use their average values for final anomaly region detection. This is achieved by selecting the top $K$ candidate regions based on their corresponding confidence scores, as shown in the following,
\begin{equation}
\label{eq:confidence-prompts}
    \mathcal{R}^C, \mathcal{S}^C := \mathrm{Top}_K(\mathcal{R}^P,\mathcal{S}^S)
\end{equation}

Denote a single region and its corresponding score as $\mathbf{r}^C$ and $s^C$, we then use these $K$ candidate regions to estimate the final anomaly map, 
\begin{equation}
\label{eq:fusion}
    \mathbf{A}_{ij} := \frac{ \sum_{\mathbf{r}^C \in \mathcal{R}^C}
    \mathbf{r}^C_{i j} \cdot s^C}{
    \sum_{\mathbf{r}^C \in \mathcal{R}^C}
    \mathbf{r}^C_{i j}}   
\end{equation}

With the proposed multimodal prompts ($ \mathcal{P}^L, \mathcal{P}^P,\mathcal{P}^S$, and $\mathcal{P}^C$), SAA is regularized and updated into our final framework, \textit{i.e.}, Segment Any Anomaly $+$ (SAA$+$), which makes more reliable anomaly predictions.

\section{Experiments}
\label{sec:experiment}
In this section, we first assess the performance of SAA/SAA$+$ on several anomaly segmentation benchmarks. Then, we extensively study the effectiveness of individual multimodal prompts.

\subsection{Experimental Setup}

\noindent\textbf{Datasets.} We leverage two datasets with pixel-level annotations: VisA~\cite{zou2022spot} and MVTec-AD~\cite{bergmann2019mvtec}, both of which comprise a variety of object subsets, \textit{e.g.}, circuit boards.

\noindent\textbf{Evaluation Metrics.} ZSAS performance is evaluated in terms of {max-F1-pixel} ($\mathcal{F}_{p}$) ~\cite{jeong2023winclip}, which measures the F1-score for pixel-wise segmentation at the optimal threshold.

\vspace{0.05in}

\noindent\textbf{Implementation Details.} We adopt the official implementations of GroundingDINO~\cite{liu2023grounding} and SAM~\cite{kirillov2023segment} to construct the vanilla baseline (SAA). Details about the prompts derived from domain expert knowledge are explained in the supplementary material. For the saliency prompts induced from image content, we utilize the WideResNet50~\cite{zagoruyko2016wideresnet} network, pre-trained on ImageNet~\cite{hinton2012imagenet}, and set $N=400$ in line with prior studies~\cite{aota2023zero}. For anomaly confidence prompts, we set the hyperparameter $K$ as $5$ by default. Input images are fixed at a resolution of $400 \times 400$.

\subsection{Main Results}
\label{sec:exp_main}

\noindent\textbf{Methods for Comparison.}  We compare our final model, \textit{i.e.}, Segment Any Anomaly + (SAA$+$) with several concurrent state-of-the-art methods, including WinClip~\cite{jeong2023winclip}, UTAD~\cite{aota2023zero}, ClipSeg~\cite{clipseg2022}, and our vanilla baseline (SAA). For WinClip, we report its official results on VisA and MVTec-AD. For the other three methods, we use official implementations and adapt them to the ZSAS task. 

\input{tables/tech_report/main_exp}

\noindent\textbf{Quantitative Results}: As is shown in Table \ref{tab:my-table}, SAA$+$ method outperforms other methods in  $\mathcal{F}_{p}$  by a significant margin. Although WinClip~\cite{jeong2023winclip}, ClipSeg~\cite{clipseg2022}, and SAA also use foundation models, SAA$+$ better unleash the capacity of foundation models and adapts them to tackle ZSAS. The remarkable performance of SAA$+$ meets the expectation to segment any anomaly without training.

\begin{figure}[t]
\vspace{-2mm}
    \centering
        \includegraphics[width=\linewidth]{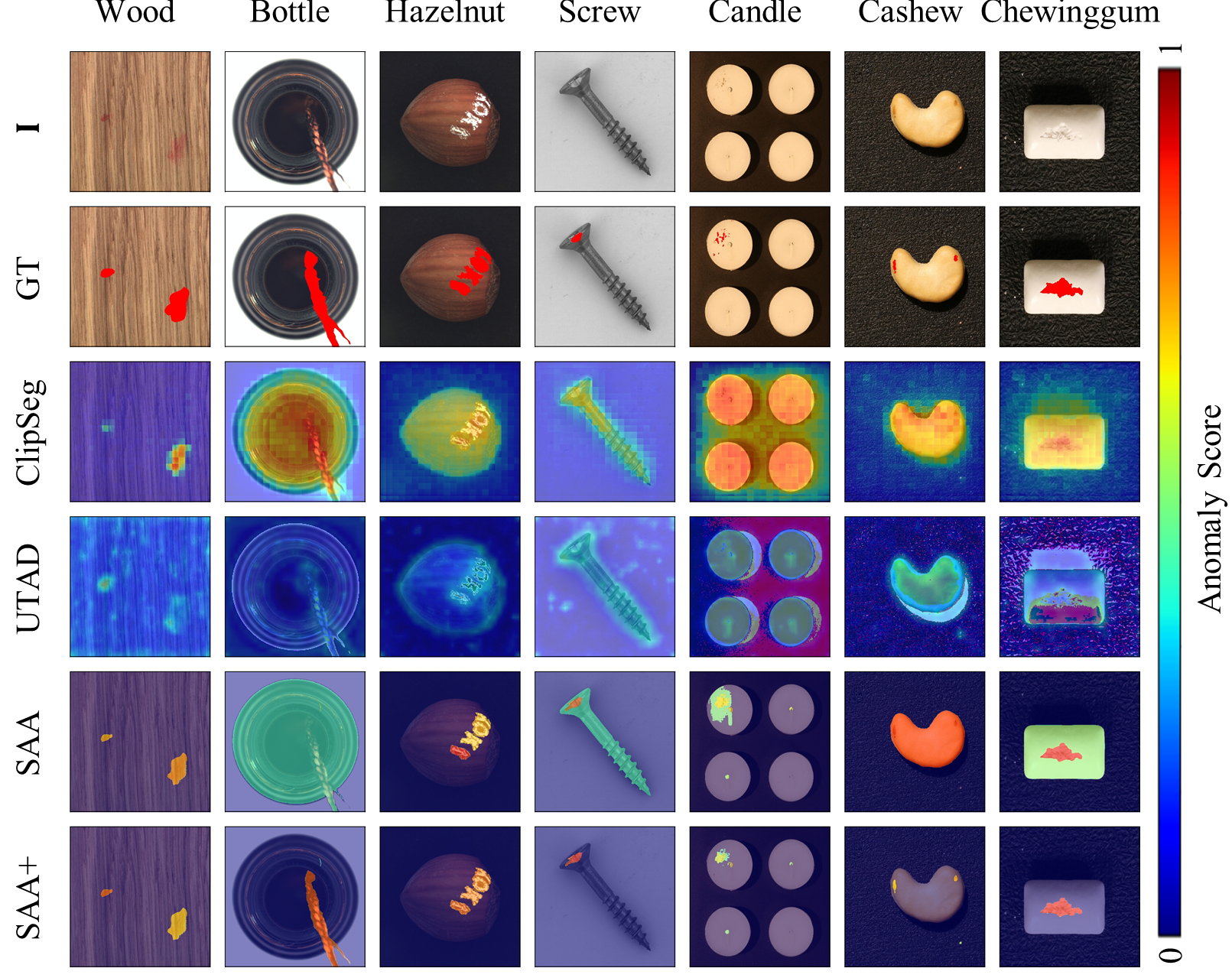}
    \vspace{-6mm}\caption{Qualitative comparisons on zero-shot anomaly segmentation  for ClipSeg~\cite{clipseg2022}, UTAD~\cite{aota2023zero}, SAA, and SAA$+$ on VisA~\cite{zou2022spot}, and MVTec-AD~\cite{bergmann2019mvtec}.}
    \label{fig:qualitative}
\end{figure}

\noindent\textbf{Qualitative Results}: Fig. \ref{fig:qualitative} presents qualitative comparisons between  SAA$+$ and previous competitive methods, where SAA$+$ achieves better performance. Moreover, the visualization shows SAA$+$ is capable of detecting all kinds of anomalies. 

\input{tables/tech_report/ablation}

\subsection{Ablation study}
\label{sec:ablation}
In Table \ref{tab:ablation}, we perform component-wise analysis to ablate specific prompt designs in our framework, which verifies the effectiveness of all the multimodal prompts, including language prompt ($\mathcal{P}^L$), property prompt ($\mathcal{P}^P$), saliency prompt ($\mathcal{P}^S$), and confidence prompt ($\mathcal{P}^C$).

\section{Conclusion}

In this work, we explore how to \textit{segment any anomaly} without any further training by unleashing the full power of modern foundation models. We owe the struggle of adapting foundation model assembly to anomaly segmentation to the prompt design, which is the key to controlling the function of off-the-shelf foundation models. Thus, we propose a novel framework, \textit{i.e.}, Segment Any Anomaly $+$, to leverage multimodal prompts derived from both expert knowledge and target image context to regularize foundation models free of training. Finally, we successfully adapt multiple foundation models to tackle zero-shot anomaly segmentation, achieving new SoTA results on several benchmarks. 


{\small
\bibliographystyle{unsrt}

}

\end{document}

%% file: tables/tech_report/main_exp.tex
\begin{table}[t]
\centering
\caption{Quantitative comparisons between SAA$+$ and other concurrent methods on zero-shot anomaly segmentation. Best scores are highlighted in \textbf{bold}.}
\label{tab:my-table}
\resizebox{0.45\textwidth}{!}{
\begin{tabular}{cccccc}
\hline
Dataset & WinClip~\cite{jeong2023winclip} & ClipSeg~\cite{clipseg2022} & UTAD~\cite{aota2023zero}  & SAA   & SAA+           \\ \hline
MVTec   & 31.65   & 25.42   & 23.48 & 23.44 & \textbf{39.40} \\
VisA    & 14.82   & 14.32   & 6.95  & 12.76 & \textbf{27.07} \\ \hline
\end{tabular}
}
\end{table}

%% file: tables/tech_report/ablation.tex
\begin{table}[t]
\centering
\caption{Ablation Study. Best scores are highlighted in \textbf{bold}.}
\label{tab:ablation}
\resizebox{0.45\textwidth}{!}{
\setlength{\tabcolsep}{16.pt}
\begin{tabular}{ccc}
\hline
{Model Variants}     & VisA           & MVTec-AD       \\ \hline
w/o $\mathcal{P}^L$                          & 23.29          & 36.49          \\
w/o $\mathcal{P}^P$                          & 19.28          & 24.43          \\
w/o $\mathcal{P}^S$                          & 19.39          & 38.79          \\
w/o $\mathcal{P}^C$                          & 26.70          & 38.68          \\
Full Model (SAA$+$)                            & \textbf{27.07} & \textbf{39.40} \\ \hline
\end{tabular}
}
\end{table}

%% file: tech_report.bbl
\begin{thebibliography}{10}

\bibitem{cao2023segment}
Yunkang Cao, Xiaohao Xu, Chen Sun, Yuqi Cheng, Zongwei Du, Liang Gao, and
  Weiming Shen.
\newblock Segment any anomaly without training via hybrid prompt
  regularization.
\newblock {\em arXiv preprint arXiv:2305.10724}, 2023.

\bibitem{liu2023grounding}
Shilong Liu, Zhaoyang Zeng, Tianhe Ren, Feng Li, Hao Zhang, Jie Yang, Chunyuan
  Li, Jianwei Yang, Hang Su, Jun Zhu, et~al.
\newblock Grounding dino: Marrying dino with grounded pre-training for open-set
  object detection.
\newblock {\em arXiv preprint arXiv:2303.05499}, 2023.

\bibitem{kirillov2023segment}
Alexander Kirillov, Eric Mintun, Nikhila Ravi, Hanzi Mao, Chloe Rolland, Laura
  Gustafson, Tete Xiao, Spencer Whitehead, Alexander~C Berg, Wan-Yen Lo, et~al.
\newblock Segment anything.
\newblock {\em arXiv preprint arXiv:2304.02643}, 2023.

\bibitem{deng2022anomaly}
Hanqiu Deng and Xingyu Li.
\newblock Anomaly detection via reverse distillation from one-class embedding.
\newblock In {\em Proceedings of the IEEE/CVF Conference on Computer Vision and
  Pattern Recognition}, pages 9737--9746, 2022.

\bibitem{cao2022informative}
Yunkang Cao, Qian Wan, Weiming Shen, and Liang Gao.
\newblock Informative knowledge distillation for image anomaly segmentation.
\newblock {\em Knowledge-Based Systems}, 248:108846, 2022.

\bibitem{wan_position_2022}
Qian Wan, Yunkang Cao, Liang Gao, Weiming Shen, and Xinyu Li.
\newblock Position encoding enhanced feature mapping for image anomaly
  detection.
\newblock In {\em 2022 {IEEE} 18th International Conference on Automation
  Science and Engineering ({CASE})}, pages 876--881. {IEEE}, 2022.

\bibitem{cao_collaborative_2023}
Yunkang Cao, Xiaohao Xu, Zhaoge Liu, and Weiming Shen.
\newblock Collaborative discrepancy optimization for reliable image anomaly
  localization.
\newblock {\em IEEE Transactions on Industrial Informatics}, pages 1--10, 2023.

\bibitem{bergmann2019mvtec}
Paul Bergmann, Michael Fauser, David Sattlegger, and Carsten Steger.
\newblock {MVTec AD} -- {A} comprehensive real-world dataset for unsupervised
  anomaly detection.
\newblock In {\em Proceedings of the IEEE/CVF conference on Computer Vision and
  Pattern Recognition}, pages 9592--9600, 2019.

\bibitem{baur_autoencoders_2021}
Christoph Baur, Stefan Denner, Benedikt Wiestler, Nassir Navab, and Shadi
  Albarqouni.
\newblock Autoencoders for unsupervised anomaly segmentation in brain mr
  images: a comparative study.
\newblock {\em Medical Image Analysis}, 69:101952, 2021.

\bibitem{radford2021learning}
Alec Radford, Jong~Wook Kim, Chris Hallacy, Aditya Ramesh, Gabriel Goh,
  Sandhini Agarwal, Girish Sastry, Amanda Askell, Pamela Mishkin, Jack Clark,
  et~al.
\newblock Learning transferable visual models from natural language
  supervision.
\newblock In {\em International Conference on Machine Learning}, pages
  8748--8763. PMLR, 2021.

\bibitem{ju_prompting_2022}
Chen Ju, Tengda Han, Kunhao Zheng, Ya~Zhang, and Weidi Xie.
\newblock Prompting visual-language models for efficient video understanding.
\newblock In {\em Computer Vision--ECCV 2022: 17th European Conference, Tel
  Aviv, Israel, October 23--27, 2022, Proceedings, Part XXXV}, pages 105--124.
  Springer, 2022.

\bibitem{jia_visual_2022}
Menglin Jia, Luming Tang, Bor-Chun Chen, Claire Cardie, Serge Belongie, Bharath
  Hariharan, and Ser-Nam Lim.
\newblock Visual prompt tuning.
\newblock In {\em Computer Vision--ECCV 2022: 17th European Conference, Tel
  Aviv, Israel, October 23--27, 2022, Proceedings, Part XXXIII}, pages
  709--727. Springer, 2022.

\bibitem{zang_unified_2022}
Yuhang Zang, Wei Li, Kaiyang Zhou, Chen Huang, and Chen~Change Loy.
\newblock Unified vision and language prompt learning.
\newblock {\em arXiv preprint arXiv:2210.07225}, 2022.

\bibitem{shen_multitask_2022}
Sheng Shen, Shijia Yang, Tianjun Zhang, Bohan Zhai, Joseph~E Gonzalez, Kurt
  Keutzer, and Trevor Darrell.
\newblock Multitask vision-language prompt tuning.
\newblock {\em arXiv preprint arXiv:2211.11720}, 2022.

\bibitem{zhou_learning_2022}
Kaiyang Zhou, Jingkang Yang, Chen~Change Loy, and Ziwei Liu.
\newblock Learning to prompt for vision-language models.
\newblock {\em Int J Comput Vis}, 130(9):2337--2348, 2022.

\bibitem{clipseg2022}
Timo L{\"u}ddecke and Alexander Ecker.
\newblock Image segmentation using text and image prompts.
\newblock In {\em Proceedings of the IEEE/CVF Conference on Computer Vision and
  Pattern Recognition}, pages 7086--7096, 2022.

\bibitem{jeong2023winclip}
Jongheon Jeong, Yang Zou, Taewan Kim, Dongqing Zhang, Avinash Ravichandran, and
  Onkar Dabeer.
\newblock Winclip: Zero-/few-shot anomaly classification and segmentation.
\newblock {\em arXiv preprint arXiv:2303.14814}, 2023.

\bibitem{object_calibration}
Xiaohao Xu, Jinglu Wang, Xiang Ming, and Yan Lu.
\newblock Towards robust video object segmentation with adaptive object
  calibration.
\newblock In {\em Proceedings of the 30th ACM International Conference on
  Multimedia}, pages 1--10, 2022.

\bibitem{xu2022reliable}
Xiaohao Xu, Jinglu Wang, Xiao Li, and Yan Lu.
\newblock Reliable propagation-correction modulation for video object
  segmentation.
\newblock In {\em Proceedings of the AAAI Conference on Artificial
  Intelligence}, pages 2946--2954, 2022.

\bibitem{paiss_count_2023}
Roni Paiss, Ariel Ephrat, Omer Tov, Shiran Zada, Inbar Mosseri, Michal Irani,
  and Tali Dekel.
\newblock Teaching clip to count to ten.
\newblock {\em arXiv preprint arXiv:2302.12066}, 2023.

\bibitem{Laion400}
Christoph Schuhmann, Robert Kaczmarczyk, Aran Komatsuzaki, Aarush Katta,
  Richard Vencu, Romain Beaumont, Jenia Jitsev, Theo Coombes, and Clayton
  Mullis.
\newblock Laion-400m: Open dataset of clip-filtered 400 million image-text
  pairs.
\newblock In {\em NeurIPS Workshop Datacentric AI}. J{\"u}lich Supercomputing
  Center, 2021.

\bibitem{zou2022spot}
Yang Zou, Jongheon Jeong, Latha Pemula, Dongqing Zhang, and Onkar Dabeer.
\newblock {SPot-the-Difference} self-supervised pre-training for anomaly
  detection and segmentation.
\newblock In {\em Proceedings of the European Conference on Computer Vision},
  2022.

\bibitem{aota2023zero}
Toshimichi Aota, Lloyd Teh~Tzer Tong, and Takayuki Okatani.
\newblock Zero-shot versus many-shot: Unsupervised texture anomaly detection.
\newblock In {\em Proceedings of the IEEE/CVF Winter Conference on Applications
  of Computer Vision}, pages 5564--5572, 2023.

\bibitem{hinton2012imagenet}
Geoffrey~E Hinton, Alex Krizhevsky, and Ilya Sutskever.
\newblock Image{N}et classification with deep convolutional neural networks.
\newblock {\em Advances in Neural Information Processing Systems},
  25(1106-1114):1, 2012.

\bibitem{zagoruyko2016wideresnet}
Sergey Zagoruyko and Nikos Komodakis.
\newblock Wide residual networks.
\newblock In Edwin R.~Hancock Richard C.~Wilson and William A.~P. Smith,
  editors, {\em Proceedings of the British Machine Vision Conference (BMVC)},
  pages 87.1--87.12. BMVA Press, September 2016.

\end{thebibliography}
